\documentclass[lettersize,journal]{IEEEtran}

\usepackage{booktabs}
\usepackage{placeins}
\usepackage{amsmath,amsfonts}
\usepackage{array}
\usepackage{float}
\usepackage{textcomp}
\usepackage{stfloats}
\usepackage{url}
\usepackage{verbatim}
\usepackage{graphicx}
\usepackage{cite}
\usepackage[bookmarks=false]{hyperref}

\usepackage{multirow}
\usepackage{graphicx} % for \graphicspath

\usepackage{caption}
\usepackage{subcaption}

\usepackage{xcolor}
\usepackage{xspace}
\usepackage{algorithm}
\usepackage[noend]{algpseudocode} % algorithmic, \State{}, etc
\usepackage{amssymb,amsmath,comment,cite}
\usepackage{ntheorem}
\newtheorem{theorem}{Theorem}

\newtheorem{lemma}{Lemma}
\newtheorem{definition}{Definition}
\newtheorem{remark}{Remark}
\usepackage{url}
\usepackage{comment}

\usepackage{ifthen}
\newboolean{shortver}
\setboolean{shortver}{false}% for long version

\newcommand\procedurename[1]{\textsl{#1}}
\newcommand\CheckForPrune{\procedurename{CheckForPrune}\xspace}

\newcommand\ComputeReachableSets{\procedurename{ComputeReachableSets}\xspace}

\newcommand{\rev}[1]{{#1}}
\newcommand\FrontierSet{\mathcal{F}}
\newcommand\abbrPF{PF}
\newcommand\abbrPFRF{\rev{GSP}\xspace}
\newcommand\abbrRAstar{$\text{RF-A}^*$\xspace}
\newcommand{\RFAnoh}{RF-A$^*_{cached}$\xspace}
\newcommand{\RFDij}{RF-A$^*_{noh}$\xspace}
\newcommand{\citep}{\cite}

\begin{document}

% \title{\LARGE\bf{Heuristic Search for Path Finding with Refuelling}}

% \author{Anushtup Nandy, Zhongqiang Ren, Sivakumar Rathinam, Howie Choset}

% \maketitle

% \thispagestyle{plain}
% \pagestyle{plain}
% \pagenumbering{arabic}

\title{Heuristic Search for Path Finding with Refuelling}

\author{Shizhe Zhao$^{1*}$, Anushtup Nandy$^{2*}$, Howie Choset$^{2}$, Sivakumar Rathinam$^{3}$ and Zhongqiang Ren$^{4}$% <-this % stops a space
	%\thanks{*This work was not supported by any organization}% <-this % stops a space
	\thanks{Manuscript received: October 15, 2024; Revised January 4, 2025; Accepted January 31, 2025}
	\thanks{This paper was recommended for publication by Editor Bera Aniket upon evaluation of the Associate Editor and Reviewers' comments.}

 \thanks{This work was supported by the National Science Foundation under Grant Nos. 2120219 and 2120529. The authors at Shanghai Jiao Tong University are supported by the Natural Science Foundation of China under Grant No. 62403313. (Corresponding author: Zhongqiang Ren.)}
 \thanks{$^{1}$ Shizhe Zhao is with UM-SJTU Joint Institute, 
 Shanghai Jiao Tong University, China. 
Email: {\tt\footnotesize shizhe.zhao@sjtu.edu.cn}}
	\thanks{$^{2}$ Anushtup Nandy and Howie Choset are at Carnegie Mellon University, 5000 Forbes Ave., Pittsburgh, PA
		15213, USA. Emails: {\tt\footnotesize \{anandy, choset\}@andrew.cmu.edu}
	}
\thanks{$^{3}$Sivakumar Rathinam is with the Department of Mechanical Engineering and the Department of Computer Science and Engineering, Texas A\&M University,
		College Station, TX 77843-3123.
		Email: {\tt\footnotesize srathinam@tamu.edu}
	}
\thanks{$^{4}$Zhongqiang Ren is with UM-SJTU Joint Institute and the Department of Automation,
 Shanghai Jiao Tong University, China. 
Email: {\tt\footnotesize zhongqiang.ren@sjtu.edu.cn}}
 \thanks{* Equal contribution and co-first authors.}
 \thanks{Digital Object Identifier (DOI): see top of this page.}
}

\markboth{IEEE ROBOTICS AND AUTOMATION LETTERS. PREPRINT VERSION. ACCEPTED January, 2025}%
{Shizhe\MakeLowercase{\textit{et al.}}: RF-A$^*$}
\maketitle

\begin{abstract}
    This paper considers a generalization of the Path Finding (\abbrPF) problem with refuelling constraints referred to as the
		\rev{Gas Station Problem (\abbrPFRF)}. Similar to PF, given a graph where vertices are gas stations with known fuel prices, and edge costs are the gas consumption between the two vertices, \abbrPFRF seeks a minimum-cost path from the start to the goal vertex for a robot with a limited gas tank and a limited number of refuelling stops.
    While \abbrPFRF is polynomial-time solvable, it remains a challenge to quickly compute an optimal solution in practice since it requires simultaneously determine the path, where to make the stops, and the amount to refuel at each stop.
    This paper develops a heuristic search algorithm called $\text{Refuel A}^*$ (\abbrRAstar) that iteratively constructs partial solution paths from the start to the goal guided by a heuristic while leveraging dominance rules for pruning during planning.
    \abbrRAstar is guaranteed to find an optimal solution and often runs 2 to 8 times faster than the existing approaches in large city maps with several hundreds of gas stations.
\end{abstract}
%%%%%%%%%%%%%%%%%%%%%%%%%%%%%%%%%%%%%%%%%%%%%%%%%%%%%%%%%%%%%%%%%%%%%%%%%%%%%%%


% \IEEEPARstart{T}{his} 

\section{Introduction}
\label{ppwr:sec:Intro}

\begin{figure}[tbp]
         \centering
         \includegraphics[width=\linewidth]{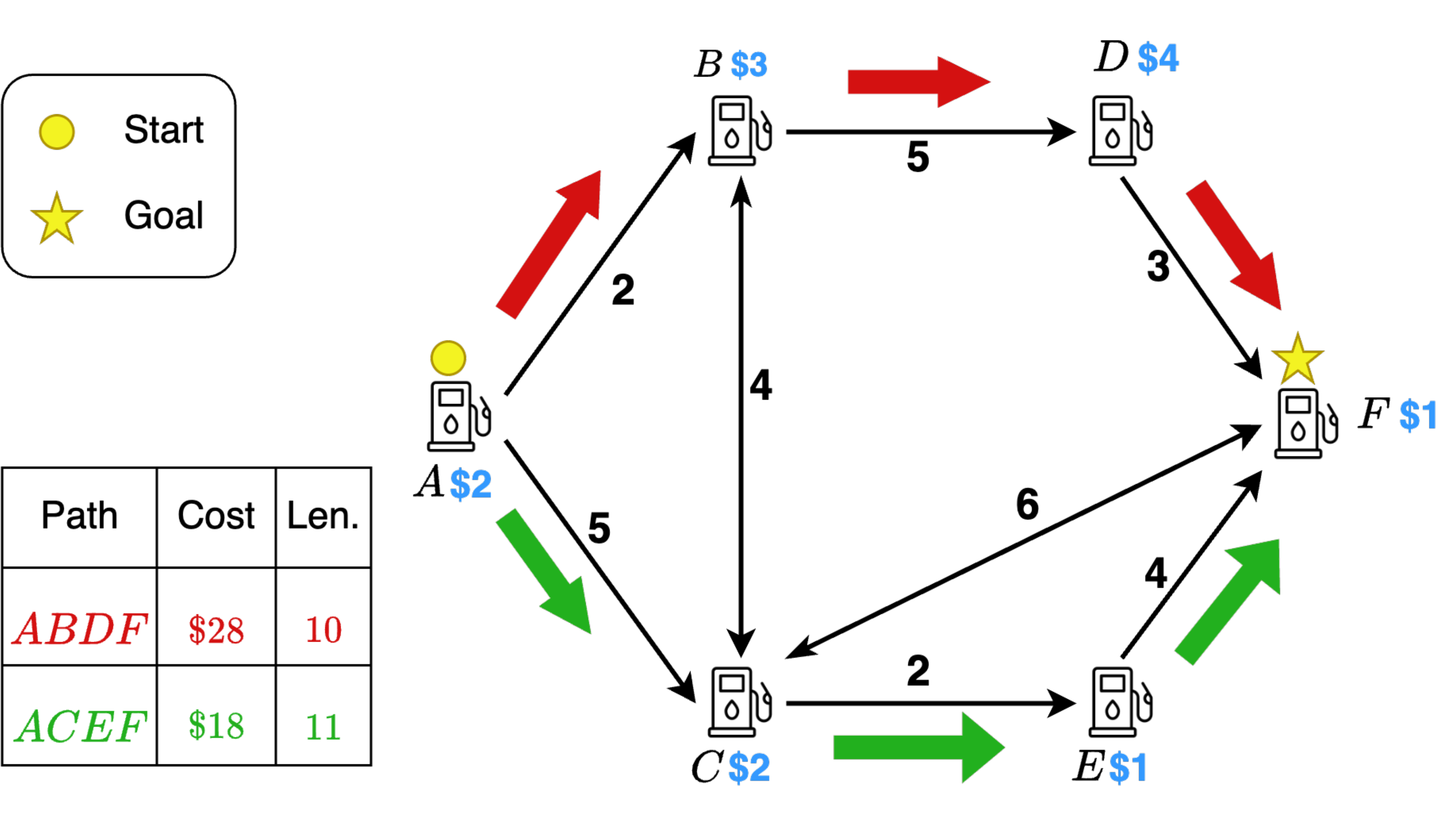}
         %\vspace{-1em}
         \caption{
         \small
         An illustrative example of \abbrPFRF. 
			 This graph consists of six vertices representing gas stations, each associated with a gas price, and each edge with its fuel expenditure. 
			The objective is to find a minimum-cost path from start to goal, 
			assuming the tank capacity is 5 and the refuelling stop limit is 3.
			The figure shows the minimum-cost path (\textit{ACEF}), using green arrows, and the minimum fuel consumption path,
			(\textit{ABDF}) using red arrows.
	Note that the minimum fuel consumption path does not incur the lowest fuel cost. 
Along the optimal solution \textit{ACEF}, the cost of refuelling at each vertex is: $\$10$ at A, $\$14$ at C, $\$18$ at E.}
         %\vspace{-2em}
         \label{fig: Graph1}
\end{figure}

\IEEEPARstart{G}{iven} a graph with non-negative edge costs, the Path Finding problem seeks a minimum-cost path from the given start vertex to a goal vertex.
This paper considers a Gas Station Problem (\abbrPFRF), where the vertices represent gas stations with known fuel prices, and the edge costs indicate the gas consumption when moving between vertices.
The fuel prices can be different at vertices and are fixed over time at each vertex.
\abbrPFRF seeks a start-goal path subject to a limited gas tank and a limited number of refuelling stops while minimizing the total fuel cost along the path (Fig.~\ref{fig: Graph1}).

\rev{\abbrPFRF was studied}~\cite{khullerFillNotFill2011,suzuki2008generic,honglinLineartimeAlgorithmFinding2007,papadopoulosFastAlgorithmGas2018}, and arises in applications such as path finding for electric vehicles between cities ~\cite{pourazarmOptimalRoutingEnergyAware2018, decauwerModelRangeEstimation2020, denunzioGeneralConstrainedOptimization2021} and package delivery using an unmanned vehicle~\cite{Lee2021, 9129495}, where a robot needs to move over long distances when refuelling becomes necessary.
While \abbrPFRF is polynomial time solvable~\cite{khullerFillNotFill2011,papadopoulosFastAlgorithmGas2018,honglinLineartimeAlgorithmFinding2007}, it remains a challenge to quickly compute an optimal solution in practice since the robot needs to simultaneously determine the path, where to make the stops, and the amount of refuelling at each stop, possibly in real-time with limited on-board computation.

This paper focuses on exact algorithms that can solve \abbrPFRF to optimality.
In~\cite{khullerFillNotFill2011}, a dynamic programming (DP) approach is developed to solve \abbrPFRF to optimality, which has been recently further improved in terms of its theoretic runtime complexity~\cite{papadopoulosFastAlgorithmGas2018}.
This approach identifies a principle regarding the amount of refuelling the robot should take at each stop along an optimal path. 
This principle \rev{allows the decomposition of \abbrPFRF into a finite number of sub-problems, and DP can be leveraged to find an optimal solution by iteratively solving all these sub-problems}.

To expedite computation, this paper develops a new heuristic search algorithm called \abbrRAstar (Refuelling A*), 
which iteratively constructs partial solution paths from the start vertex to the goal guided by a heuristic function.
\abbrRAstar gains computational benefits over DP in the following aspects.
First, \abbrRAstar never explicitly \rev{explores all sub-problems} as DP does and only explores the \rev{sub-problems} that are needed for the search.
Second, \abbrRAstar uses a heuristic to guide the search, reducing the number of \rev{sub-problems} to be explored before an optimal solution is found.
Third, taking advantage of our prior work in multi-objective search~\cite{renERCANewApproach,ren22emoa,ren2025emoa}, \abbrRAstar introduces a dominance rule to prune partial solutions during the search, which saves computation.
\abbrRAstar is guaranteed to find an optimal solution.

We compare \abbrRAstar against DP~\cite{khullerFillNotFill2011} and a method based on mixed-integer programming (MIP)~\cite{liaoElectricVehicleTouring2016} in real-world city maps of various sizes from the OpenStreetMap dataset.
Our results show that \abbrRAstar is orders of magnitude faster than the MIP, and is often 2 to 8 times faster than DP in the city maps with hundreds of gas stations.
In addition, \abbrRAstar is up to 64 times faster than DP, when the heuristic can be pre-computed and cached for planning.
%Our results show that \abbrRAstar runs more than an order of magnitude faster than the DP method as tested in maps with hundreds of gas stations.
%While DP takes up to hundreds of seconds to solve these test instances, \abbrRAstar often takes less than a second.
\rev{These results demonstrate the scalability of \abbrRAstar,
enabling it to plan for a robot with a limited tank in large urban areas.}
% Our software will be open-sourced.
% to benefit the community.\footnote{\url{https://github.com/wonderren/dev_erca_refill}}

\section{Related Work}\label{sec:related-work}

Path planning with refueling constraints (\abbrPFRF) involves determining a path and a refueling schedule simultaneously.
Dynamic programming methods were introduced by \cite{khullerFillNotFill2011,papadopoulosFastAlgorithmGas2018}
for the \abbrPFRF either from a given start to a goal or for all-pair vertices in the graph.
Besides planning start-goal paths, another related problem generalizes travelling salesman problem and
vehicle routing problem with refuelling
constraints~\cite{khullerFillNotFill2011,liaoElectricVehicleTouring2016,swedaOptimalRechargingPolicies2017,DBLP:journals/cor/KarakostasS24,
DBLP:journals/tase/AlyassiKKCET23}.
These problems seek a tour that visits multiple vertices subject to a limited fuel tank.
This paper only considers finding a start-goal path.

\rev{
Recently, due to the prevalence of electric vehicles (EV) and unmanned aerial vehicles (UAV), 
several variants of \abbrPFRF were proposed to address fuel constraints using various methods.
Mathematical programming models were proposed by~\cite{DBLP:conf/ipco/PulyassaryKSSW24}, 
which generalized the refuelling cost in EV applications. 
\cite{liOnlineRoutingAutonomous2022} addresses the refuelling constraints for an UAV cruise system under an online
setting using a greedy approach.
\cite{DBLP:journals/corr/abs-2008-03333} considers the fairness of resource utilization and
employs \abbrPFRF as a cost function. 
\cite{DBLP:journals/transci/SwedaDK17a} considers the uncertainty of the waiting time for refuelling and proposed a
policy that can be solved by dynamic programming. 
Other methods also appear in literature to deal with refuelling constraints, such as constraint
programming\cite{DBLP:journals/jirs/RamasamyRDCB22} and learning \cite{ottoniReinforcementLearningTraveling2022,liOnlineRoutingAutonomous2022}.

Among these methods, 
both mathematical programming and dynamic programming can guarantee solution optimality for \abbrPFRF.
Mathematical programming models the \abbrPFRF as a mixed integer program and then 
invokes an off-the-shelf solver to handle the problem~\cite{DBLP:conf/ipco/PulyassaryKSSW24}.
Dynamic programming (DP) decomposes the \abbrPFRF to a finite number of sub-problems, 
then exploits the relation between these sub-problems to find the optimal solution by solving all sub-problems
iteratively~\cite{khullerFillNotFill2011}. 
However, DP often performs redundant work.
This paper aims to reduce the redundancy of DP in \abbrPFRF by heuristic search.
}

\section{Problem Statement}
\label{ppwr:sec:problem_state}
    Let $G=(V,E)$ denote a directed graph, where each vertex $v\in V$ represents a gas station, and each edge $(u,v) \in E$ denotes an action that transits the robot from vertex $u$ to $v$.
    Each edge $(u,v)\in E$ is associated with a non-negative real value $d(u,v) \in \mathbb{R}^+$, called edge cost, which represents the
		amount of fuel needed to traverse the edge from $u$ to $v$.
    The robot has a fuel capacity $q_{max}\in \mathbb{R}^+$ representing the maximum amount of fuel it can store in its tank.
    Let $c: V\rightarrow [0,\infty]$ denote the refuelling price per unit of fuel at each vertex in $v \in V$.\footnote{For vertices $v$ in $G$ where the robot cannot refuel, let $c(v)=\infty$.}
    
    Let a path $\pi(v_1,v_\ell)=(v_1,v_2,\dots,v_\ell)$ be an ordered list of vertices in $G$ such that every pair of
		adjacent vertices in $\pi(v_1,v_\ell)$ is connected by an edge in $G$, i.e., \rev{$(v_i,v_{i+1}) \in E,
		i=1,2,\dots,\ell-1$}.
    Let $g(\pi)$ denote total fuel cost along the path; specifically, let a non-negative real number $a(v) \in
		\mathbb{R}^+$ denote the amount of refuelling taken by the robot at vertex $v$, then
		\rev{$g(\pi)=\sum_{i=1,2,\dots,\ell} a(v_k)c(v_i)$}.
    
    In practice, the robot often has to stop to refuel, which slows down the entire path execution time. Therefore, let $k_{max} \in \mathbb{Z}^{+}, k_{max}>1$ denote the maximum number of refuelling stops the robot is allowed to make along its path.
    
		\begin{definition}[\rev{Gas Station Problem (\abbrPFRF)}~\cite{khullerFillNotFill2011}]\label{ppwr:def:problem}
    Given a pair of start and goal vertex $v_o, v_g \in V$, the robot has zero amount of fuel at $v_o$ and must refuel to travel. 
	\abbrPFRF seeks a path $\pi$ from $v_o$ to $v_g$ and the amount of refueling $a(v), v\in\pi$ along the path, such that $g(\pi)$ is minimized, while the number of refuelling stops along $\pi$ is no larger than $k_{max}$,
		\rev{and the amount of fuel in the tank is not greater than $q_{max}$ after each refuelling}.
    \end{definition}

% We further consider a variant of the PPwR problem where the robot can make a maximum of $M$ stops for refuelling along its path before reaching the goal vertex.

% \begin{definition}[Constrained Path Planning with Refueling (CPPwR)]
% Given a pair of start and goal vertex $v_s, v_g \in V$, the robot starts with zero unit of fuel at $v_s$ and can refuel for at most $M$ times, the CPPwR problem seeks to find a path $\pi$ for the robot from $v_s$ to $v_g$ such that $g(\pi)$ reaches the minimum.
% \end{definition}

% \begin{remark}
% In Def.~\ref{ppwr:def:problem}, we only need to consider the case where the robot starts with zero fuel at $v_o$ for the following reason.
% If the robot starts with $q_0$ fuel at $v_o$, one can always construct a new problem where the robot starts with zero fuel as follows.
% First, let $G'$ denote a new graph whose vertex set is $V'=V\cup \{v_o'\}$, where $v_o'$ is only connected to $v_o$ with $d(v_o',v_o)=q_{max}-q_0$ and $c(v_o')=0$.
% Then, in $G'$, the robot starts with zero fuel, and the goal is to find a minimum cost path $\pi'$ in $G'$ from $v_o'$ to $v_g$.
% Following $\pi'$, the robot arrives at $v_o$ with $q_0$ fuel, and the remaining path is the desired solution.
% %Furthermore, since the robot starts with zero fuel, $k_{max}$ must be greater than one to refuel at $v_s$ to start its motion.
% We can also assume $c(v_g)=0$ since in an optimal solution, the robot never refuels at the goal vertex and taking $c(v_g)=0$ does not change an optimal solution.
% \end{remark}

\begin{remark}
% \shizhe{min-cost is undefined and ambiguous}
    In the literature~\cite{khullerFillNotFill2011,liaoElectricVehicleTouring2016}, the graph is often assumed to be a fully connected graph.
    For a graph $G$ that is not fully connected, one can convert it to a fully connected graph $G'$ by finding a minimum edge cost path $\pi(u,v)$ (without any limit on the number of refuel stops) for each pair of vertices $(u,v)$ in $G$ and using an edge $e' \in G'$ to indicate path $\pi(u,v)$.
		Although such a conversion is straightforward, it may \rev{require extra runtime} when deploying the algorithm on a robot.
    For this reason, this paper does not assume the graph $G$ is fully connected and presents \abbrPFRF on general graphs.
\end{remark}

%%%%%%%%%%%%%%%%%%%%%%%%%%%%%%%%%%%%%%%%%%%%%%%%%%%%%%%

% \input{MIP}

%%%%%%%%%%%%%%%%%%%%%%%%%%%%%%%%%%%%%%%%%%%%%%%%%%%%%%%

% \input{greedy}

%%%%%%%%%%%%%%%%%%%%%%%%%%%%%%%%%%%%%%%%%%%%%%%%%%%%%%%
\section{Method}
\label{ppwr:sec:Method}

\label{ppwr:sec:Method:intuitive_overview}
This section introduces \abbrRAstar, a heuristic search approach to find an optimal solution for \abbrPFRF.
It initiates at start vertex $v_o$ and systematically explores potential paths from $v_o$ towards 
goal vertex $v_g$ while minimizing the overall fuel cost.
The heuristics help estimate the remaining cost to the goal and guide the search.
\abbrRAstar also considers the fuel tank limit $q_{max}$ and the refuelling stop limit $k_{max}$, by comparing two paths that reach the same vertex using multiple criteria.
During the search, \abbrRAstar maintains an open set of candidate paths that are to be expanded, similar to A$^*$.
It continues searching until finding an optimal path satisfying the constraints.
% Our algorithm employs efficient exploration with heuristic search to solve the \abbrPFRF adeptly.
A toy example of the search process is provided in Fig.~\ref{fig:working}.

\subsection{Notations and Background}
\label{ppwr:sec:Method:Notation}

\subsubsection{Basic Concepts}
In \abbrPFRF, there can be multiple paths from $v_o$ to a vertex $v$, and to differentiate them, we use the notion of labels.
% \footnote{Other names such as states~\cite{ren22mopbd} are also used to identify paths during the search. We use label in this paper and reserve the term state for the dynamic programming method in~\cite{khullerFillNotFill2011} that is summarized in Sec.~\ref{ppwr:sec:Method:dp_baseline}.}
Intuitively, a label $l=(v,g,q,k)$
%represents a sub-problem ($v, q, k$) and the corresponding cost $g$. Specifically, a label 
consists of a vertex $v \in V$, a non-negative real number $g \in \mathbb{R}^{+}$ that represents
the cost-to-come from $v_o$ to $v$, a non-negative real number $q\in \mathbb{R}^{+}$ that represents the amount of fuel
remaining at $v$ before refuelling, \rev{and an integer $0\leq k< k_{max}$ indicates the number of refuelling
stops before $v$}.
We use $v(l),g(l),q(l),k(l)$ to denote the respective component of a label.
% Each label $l$ represents a path from $v_0$ to $v$ and has the components $v(l)$ representing the vertex, with cost $g(l)$, $q(l)$ fuel remaining at $v$ before refuelling, and total stops made $k(l)$.
To compare labels, we use the following notion of label dominance.

\begin{definition}\label{ppwr:def:dominance}
    Given two labels $l,l'$ with $v(l)=v(l')$, label $l$ dominates $l'$ if the following three inequalities hold: (i) $g(l) \leq g(l')$ ,  (ii) $q(l) \geq q(l')$ and  
    (iii) $k(l) \leq k(l')$.
\end{definition}

If $l$ dominates $l'$, then $l'$ can be discarded during the search, since for any path from $v_o$ via $l'$ to $v_g$, there must be a corresponding path from $v_o$ via $l$ to $v_g$ with the same or smaller cost.
Otherwise, both $l$ and $l'$ are non-dominated by each other.
\rev{For a vertex $v\in V$, let $\FrontierSet(v)$ denote a set of labels that reach $v$ and are non-dominated by each other.} 
$\FrontierSet(v)$ is also called the \emph{frontier set} at $v$.
% This set consists of labels that reach $v$ and are non-dominated by each other.
Additionally, the procedure $\CheckForPrune(l)$ compares \rev{a label} $l$ against all existing labels in $\FrontierSet(v(l))$ to check if $l$ is dominated and should be discarded.
% Then add $l$ to $\FrontierSet(v(l))$.

Similarly to A$^*$, let $h(l)$ denote the $h$-value of label $l$ that estimates the cost-to-go from $v$ to $v_g$.
We further explain the heuristic in Sec.~\ref{ppwr:sec:Method:heuristic_det}.
Let $f(l)=g(l)+h(l)$ be the $f$-value of label $l$.
Let OPEN denote a priority queue of labels, where labels are prioritized based on their $f$-values from the minimum to the maximum.
% Let $\FrontierSet(v),v\in V$ denote the frontier set at vertex $v$, which stores the labels that reach $v$ and are expanded during the search.

A major difficulty in \abbrPFRF is determining the amount of refuelling at each vertex during the search, a continuous variable that can take any value in $[0, q_{max}]$.
To handle this difficulty, we borrow the following lemma from~\cite{khullerFillNotFill2011}, which provides an optimal strategy for refuelling at any vertex.

\begin{lemma}[Optimal Refuelling Strategy]
\label{ppwr:lemma1}
  Given refuelling stops $v_1, \dots, v_n$ along an optimal path using at most $k_{max}$ stops \rev{in a complete graph}. 
	At $v_{g-1}$, which is the stop right before the goal vertex $v_g$, refuel enough to reach $v_g$ with an empty tank. Then, an optimal strategy to decide how much to refuel at each stop for any $n < g-1$ :
    \begin{itemize}
        \item if $c(v_n) < c(v_{n+1})$, then fill up entirely at $v_{n}$.
        \item if $c(v_n) \geq c(v_{n+1})$, then fill up enough to reach $v_{n+1}$.
    \end{itemize}
\end{lemma}

The intuition behind Lemma~\ref{ppwr:lemma1} is that the robot either fills up the tank if the next stop has a higher fuel price, or fills just enough amount of fuel to reach the next stop if the next stop has a lower price.
By doing so, the robot minimizes its accumulative fuel cost.
A detailed proof is given in \cite{khullerFillNotFill2011}.
\rev{Note that Lemma~\ref{ppwr:lemma1} assumes the graph is complete. 
To adapt it to a general graph, we need to identify all possible transitions from one vertex to any other vertices in $G$ (going through one or multiple edges) without refuelling,
which will be described in \emph{ComputeReachableSets} in the next section.}

% We introduce the following notion of label dominance to compare two labels $l,l'$ with $v(l)=v(l')$.

% \begin{definition}\label{ppwr:def:dominance}
%     Given two labels $l,l'$ with $v(l)=v(l')$, label $l$ dominates $l'$ if the following three inequalities hold: (i) $g(l) \leq g(l')$ ,  (ii) $q(l) \geq q(l')$ and  
%     (iii) $k(l) \leq k(l')$.
% \end{definition}

% If $l$ dominates $l'$, then $l'$ can be discarded during the search.
% Otherwise, $l$ and $l'$ must be stored in $\FrontierSet(v(l))$ during the search.
% $\CheckForPrune(l)$ compares $l$ against all existing labels in $\FrontierSet(v(l))$.
% Then add $l$ to $\FrontierSet(v(l))$.

\begin{algorithm} [t]
        \small
\caption{\abbrRAstar}\label{ppwr:alg:refuelA}
	\begin{algorithmic}[1]
        \State{\textsl{ComputeReachableSets}()}
        \State{\textsl{ComputeHeuristic}($v_g$)}
		\State{$l_o \gets (v_o, g=0, q=0, k=0)$, ${f}(l_o) \gets {0} + {h}(l_o)$}
            \State{$parent(l_o)\gets NULL$}\label{erca:alg:erca:line:initialLabel}
		\State{Add $l_o$ to OPEN}
		\State{$\FrontierSet(v)\gets \emptyset, \forall v \in V$}
		\While{OPEN $\neq \emptyset$}
		\State{pop $l=(v,g,q,k)$ from OPEN}\label{alg:whileBegin}
		\If{\CheckForPrune($l$,$\FrontierSet(v(l))$)}\label{erca:alg:erca:line:checkForPrune}
		\State{\textbf{continue}}
		\EndIf
		\State{add $l$ to $\FrontierSet(v(l))$}\label{erca:alg:erca:line:filterAndAddFront}
		\If{$v(l) = v_g$}\label{erca:alg:erca:line:vdCheck}
            \State{\textbf{continue}}
            \EndIf
% 		\State{Add $l$ to $\mathcal{S}$}
            \If{$k = k_{max}$}\label{alg1:line:kmaxCheck}
            \State{\textbf{continue}}
            \EndIf
		%\State{\textbf{break}}\label{erca:alg:erca:line:break}
		\ForAll{$v' \in$ \textsl{GetReachableSet}($v(l)$)}\label{erca:alg:erca:line:getSucc}
            \If{$c(v') > c(v)$}\label{alg1:line:gqk1}
            \State{$g' \gets g(l) + (q_{max} - q(l)) c(v)$}
            \State{$q' \gets q_{max} - d(v,v')$}
            \State{$k' \gets k + 1$}
            \Else
            \If{$d(v',v) \geq q(l)$}
            \State{$g' \gets g(l) + (d(v',v) - q(l)) c(v)$}
            \State{$q' \gets 0$}
            \State{$k' \gets k + 1$}\label{alg1:line:gqk2}
            \Else
            \State{\textbf{continue}}\label{alg:prune}\Comment{No need to refuel}
            % \State{$g' \gets g(l)$ (This can be skipped, TODO) }
            % \State{$q' \gets q(l)-d(v',v)$}
            % \State{$k' \gets k $}
            \EndIf
            \EndIf
		\State{$l' \gets (v', g', q', k')$}
            \State{$g(l') \gets g'$}
            \If{\CheckForPrune($l'$,$\FrontierSet(v(l))$)}
            \State{\textbf{continue}}
            \EndIf
		\State{${f}(l') \gets {g}(l') + {h}(v(l'))$}\label{erca:alg:erca:line:newLabel}
            \State{$parent(l')\gets l$}\label{erca:alg:erca:line:setParent}
            \State{add $l'$ to OPEN}\label{alg:whileEnd}
		\EndFor
		\EndWhile
		\State{\textbf{return} \textsl{Reconstruct}($v_d$)}
	\end{algorithmic}
\end{algorithm}

\begin{algorithm}[t]
\caption{\textsl{ComputeReachableSets}}\label{ppwr:alg:preG}
	\begin{algorithmic}[1]
        \small
        \State{$Reach(v) \gets \emptyset, \forall v\in V$}
        \For{$v\in V$}
				\State{$d^*(u) \gets \infty, \forall u\in V$}\label{alg2:begin}
        \State{$d^*(v) \gets 0$}
        \State{Add $v$ to OPEN$_{v}$}
        \While{OPEN$_{v}$ $\neq \emptyset$}
        \State{pop $u$ from OPEN$_{v}$}
        \If{$d^*(u) > q_{max}$}
        \State{\textbf{continue}}
        \Else
        \State{add $u$ to $Reach(v)$}
        \EndIf
        \For{$u' \in$ \textsl{GetSucc}($u$)}
        \If{$u' \in Reach(v)$}
        \State{\textbf{continue}}
        \EndIf
        \If{$d^*(u') > d^*(u) + d(u,u')$}
        \State{$d^*(u') \gets d^*(u) + d(u,u')$}
				\State{add $u'$ to OPEN$_{v}$}\label{alg2:end}
        \EndIf
        \EndFor
        \EndWhile
        \EndFor
	\end{algorithmic}
\end{algorithm}

\begin{algorithm} [t]
    \caption{\textit{$\CheckForPrune(l, \FrontierSet(v(l)))$}}
    \label{ppwr:alg:checkPrune}
        \small
    \begin{algorithmic}[1]
    \State{INPUT: A label $l$ and $\FrontierSet(v(l))$, the frontier set at vertex $v(l)$.}
    \ForAll{$l' \in \FrontierSet(v(l))$}
        \If{$g(l') \leq g(l)$ \text{and} { $q(l') \geq q(l)$} \text{and} {$k(l') \leq k(l)$} }
        \label{alg3:line:DomCheck}
            \State{\textbf{return} true} \Comment{$l$ should be pruned.}
        \EndIf
    \EndFor
    \State{\textbf{return} false}\Comment{$l$ should not be pruned.}
    \end{algorithmic}
\end{algorithm}

%%%%%%
% Given a label l, check if l is dominated by any existing label l' when v(l) = v(l').
% g(l), q(l); g(l'), q(l'); c(v(l)) - gas price at this vertex.
% g(l') + (q(l)-q(l'))*c(v(l)) <? g(l), if yes, then l can be discarded.
% Otherwise, l should be kept.

% \begin{algorithm} [htbp]
%     \caption{\textit{$\CheckForPrune(l, \FrontierSet(v(l)))$}} \label{ppwr:alg:checkPrune}
%     \begin{algorithmic}[1]
%     \State{INPUT: A label $l$ and $\FrontierSet(v(l))$, the frontier set at vertex $v(l)$.}
%     \ForAll{$l' \in \FrontierSet(v(l))$}
%         \If{$g(l') + (q(l) - q(l'))*c(v) \leq g(l)$}
%             \State{\textbf{return} true} \Comment{$l$ should be pruned.}
%         \EndIf
%     \EndFor
%     \State{\textbf{return} false}\Comment{$l$ should not be pruned.}
%     \end{algorithmic}
% \end{algorithm}

\subsection{$\text{Refuel A}^{*} \text{ Algorithm}$}
\label{ppwr:sec:Method:refuel-alg}

\abbrRAstar (Alg.~\ref{ppwr:alg:refuelA}) takes \rev{a graph} $G$, \rev{tank capacity} $q_{max}$, $v_o$, $v_g$, and
\rev{max refuelling stops} $k_{max}$ as the inputs. It begins by calling Alg.~\ref{ppwr:alg:preG} \ComputeReachableSets (\ref{ppwr:sec:Method:checkprune}) to compute the set of all vertices that the robot can travel to from any vertex $u$ given a full tank.
% \footnote{Computing reachable sets can be avoided if the given graph $G$ is fully connected.}
Subsequently, to compute the heuristic which gives the amount of fuel needed to reach $v_g$ from any other vertex, it runs an exhaustive backward Dijkstra search from $v_g$.
We present the heuristic computation in Sec.~\ref{ppwr:sec:Method:heuristic_det}.
After the Dijkstra, \abbrRAstar initiates the label $l_o = (v_o, g=0, q=0, k=0)$ at vertex $v_o$ with the ${f}$-value, ${f}(l_o) = {h}(l_o)$, and inserts it into $\text{OPEN}$. The frontier set $\FrontierSet(v)$ at each $v\in V$ is initialized as an empty set $\emptyset$. 

During the search (Lines \ref{alg:whileBegin}-\ref{alg:whileEnd}), in each iteration, the label with the lowest f-value is {popped} from OPEN for further processing. This label is checked for dominance against existing labels in $\FrontierSet(v(l))$ using $\CheckForPrune$. This procedure employs Def.~\ref{ppwr:def:dominance} to compare the $g$, $q$ and $k$ values of the popped label against other labels in $\FrontierSet(v(l))$.

If the selected label is non-dominated (and thus unpruned), it is added to the frontier set. Subsequently, the algorithm
checks if \rev{the vertex of the label} $v(l)$ is $v_g$, which means that the label $l$ represents a solution with the minimum cost, and the search terminates.
It is also confirmed whether the $k_{max}$ stops limit has been reached. In cases where $v(l) \neq v_g$ and $k' \neq k_{max}$, the label is expanded, which generates new labels for all reachable vertices from $v(l)$ in $G$.
This involves a loop that iterates each reachable vertex and creates a new label $l'$ with new $g', q', k'$.
The amount of refuelling is determined using Lemma~\ref{ppwr:lemma1}, and the corresponding accumulative fuel cost $g'$ is computed. 
Note that $l'$ is generated only if a refuelling stop at $v(l)$ is required.
$v'$ at Line~\ref{alg:prune} can thus be skipped.
Finally, the algorithm uses $\CheckForPrune$ to check for dominance. If $l'$ is not pruned, $l'$ is added to OPEN for future expansion.

\subsubsection{\ComputeReachableSets}
\label{ppwr:sec:Method:compute_reach}

This procedure aims to find all successor vertices that \abbrRAstar needs to consider when expanding a label.
To achieve this, it identifies all vertices $v' \in V$ that the robot with a full tank can reach from vertex $v$ \rev{without refuelling}.
Specifically, Alg.~\ref{ppwr:alg:preG}, initializes $\text{Reach}(v)$ as an empty set for each $v \in V$.
In each iteration, it designates a vertex $v$ and runs a Dijkstra search from $v$ to all other vertices in $G$ to find vertices that are reachable from $v$ without refuelling.
Lines~\ref{alg2:begin}-\ref{alg2:end} show this Dijkstra search process starting from a specific vertex $v$.
This algorithm involves iterating through vertices in $V$ and traversing their successors to update the distances.% hence it has a run-time complexity of $O(|E|\log |V| + |V|\log |V|)$.
% \footnote{Here we use a "set" to implement the OPEN set.} 
%{ \red Dijkstra's complexity depends on how OPEN (a heap or say priority queue) is implemented. I think you need to add a footnote mentioning what type of priority queue implementation you are assuming.}

\subsubsection{Heuristic Computation}
\label{ppwr:sec:Method:heuristic_det}

A possible way to compute the heuristic is to first run an exhaustive backward Dijkstra search on $G$ from $v_g$ to any
other vertices in $G$ using \rev{fuel consumption} $d(u,v)$ (ignoring the fuel tank limit of the robot and the refuelling cost).
After this Dijkstra search, let $d_{v_g}(v)$ denote the minimum fuel consumption path from $v$ to $v_g$.
Let $c_{min} := \min_{v\in V - \{v_g\}}( c(v))$ denote the minimum fuel price in $G$.
Then, let $h(l) = \max \{(d_{v_g}(v(l))-q(l)) c_{min}, \;0\}$ be the $h$-value of label $l$. 
% The admissibility of the heuristic is proved through Lemma ~\ref{ppwr:lemma5}.
When computing this heuristic, the tank limit of the robot is ignored and the fuel price at any station is a lower bound of the true price at that station.
As a result, $h(v)$ provides a lower bound of the total fuel cost to reach $v_g$ from $v$.
We therefore have the following lemma.
\begin{lemma}[Admissible Heuristic]
\label{ppwr:lemma:admissibleHeu}
    The heuristic, $h(l) = max\{(d_{v_g}(v(l)) - q(l))c_{min}, 0\}$, is admissible.
\end{lemma}

\subsubsection{$\CheckForPrune(l, v(l))$}
\label{ppwr:sec:Method:checkprune}

As shown in Alg.~\ref{ppwr:alg:checkPrune}, this procedure checks for dominance for each label $l$
against all labels in $\FrontierSet(v(l))$, \rev{the frontier set at vertex $v(l)$}, by using Def.~\ref{ppwr:def:dominance}. Alg.~\ref{ppwr:alg:checkPrune} iterates the frontier set $\FrontierSet$, with its efficiency determined by the size $|\FrontierSet(v(l))|$, leading to a complexity of $O(|\FrontierSet(v(l))|)$.
This dominance check can be expedited by the techniques in~\cite{ren22emoa,renERCANewApproach}, which is left for our future work.

\begin{remark}\label{remark:empty-tank}
	We use the same definition of \abbrPFRF as in~\cite{khullerFillNotFill2011}, 
	which assumes the robot always starts with an empty tank. 
	%This allows Line~\ref{alg:prune} of Alg~\ref{ppwr:alg:refuelA} to be safely pruned,
	%as we can always reduce the amount of refueling at a previous stop to achieve a lower cost.
	%However, this reasoning no longer holds when the initial tank is not empty.
	For the cases where the initial tank at $v_o$ is not empty (i.e, $q_o>0$), 
	one can construct a new problem that starts with an empty tank at a pseudo vertex $v_p$, and $v_p$ is only connected with $v_o$ such that, by following the Lemma~\ref{ppwr:lemma1}, the robot can only fill up the tank at $v_p$ and then reaches $v_o$ with $q_o$ amount of remaining fuel.
	More details about constructing the new problem is described in~\cite{khullerFillNotFill2011}.
\end{remark}

% \begin{figure}[tb]
%   \centering
%   \includegraphics[width=\linewidth]{Pics/Example_Refuel_A.jpeg} 
%   \caption{
%   \small
%   A toy example showing the $\text{Refuel A}^{*}$ with four vertices. (a) The entire graph has four vertices $(v,u,b \text{ and } w)$, with fuel cost in red, fuel spent in blue, $q_{max}=6$ and $k_{max}=2$. The labels associated with each vertex are $l_{vertex} = (u, q, g, k)$. (b) The first iteration in the search case is where $l_o$ is popped from OPEN. Around each vertex, its corresponding label is provided. (c) The next iteration in the search is provided, where $l_2$ is popped, and (d) shows the last case when $l_5$ is popped. Optimal path is $v \to u \to b$}
%   \label{fig:working}
%   % \vspace{-1.5em}
% \end{figure}

\begin{figure}[tb]
	\centering
	\begin{subfigure}{.50\linewidth}
		\includegraphics[width=\linewidth]{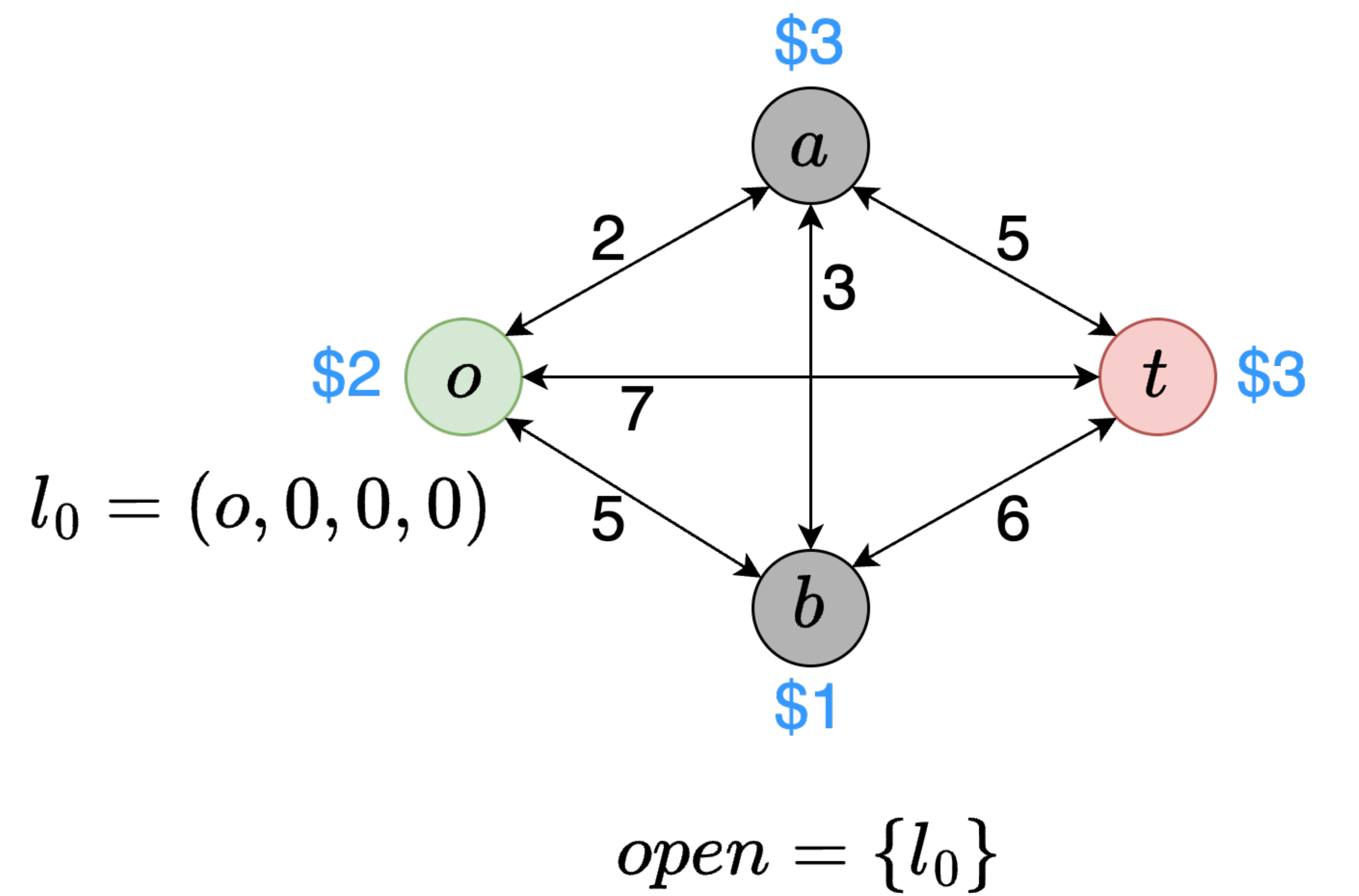}	
		\caption{}
		\label{fig:rf-demo-1}
	\end{subfigure}%
	\begin{subfigure}{.50\linewidth}
		\includegraphics[width=\linewidth]{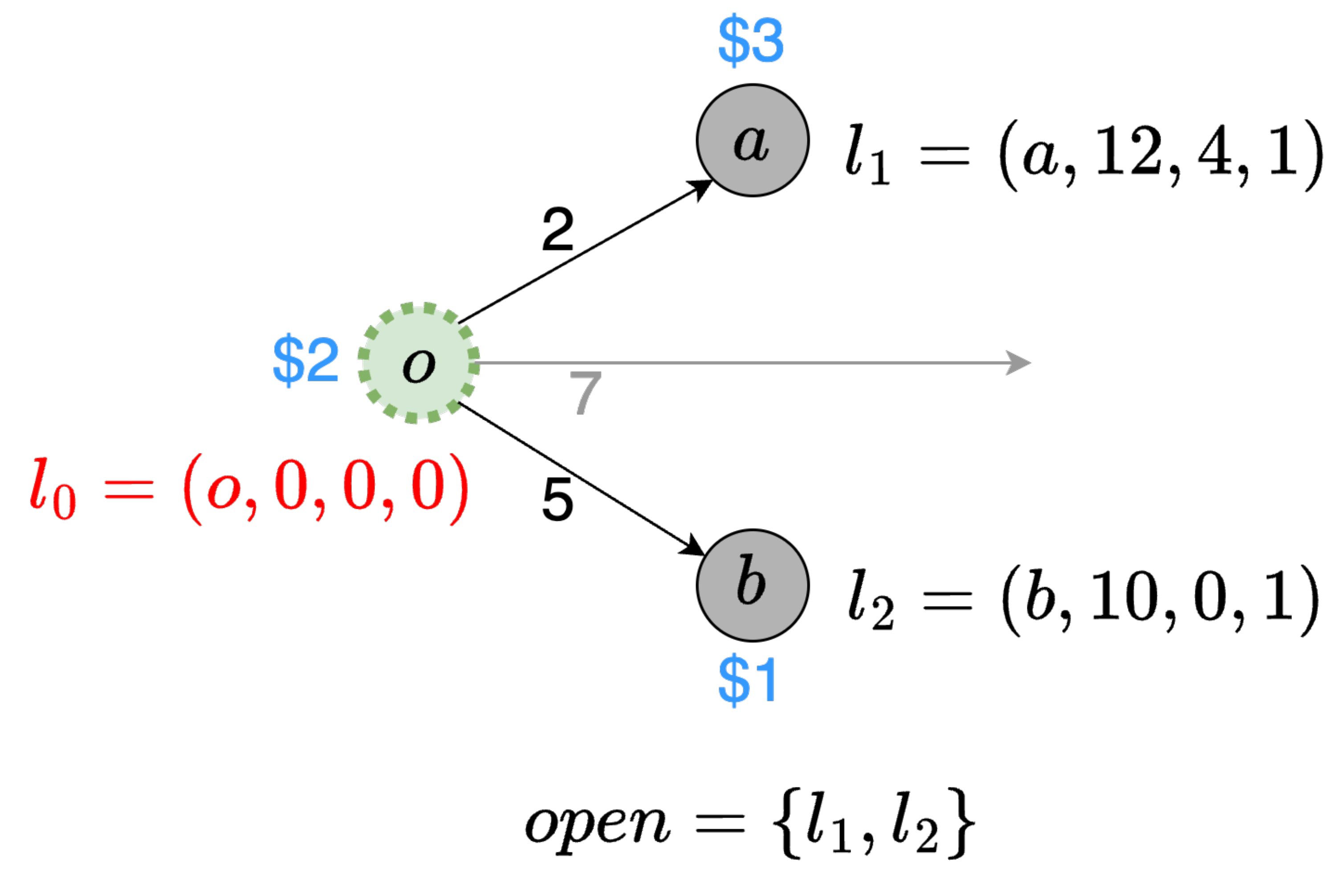}	
		\caption{}
		\label{fig:rf-demo-2}
	\end{subfigure}
 
	\begin{subfigure}{.50\linewidth}
		\includegraphics[width=\linewidth]{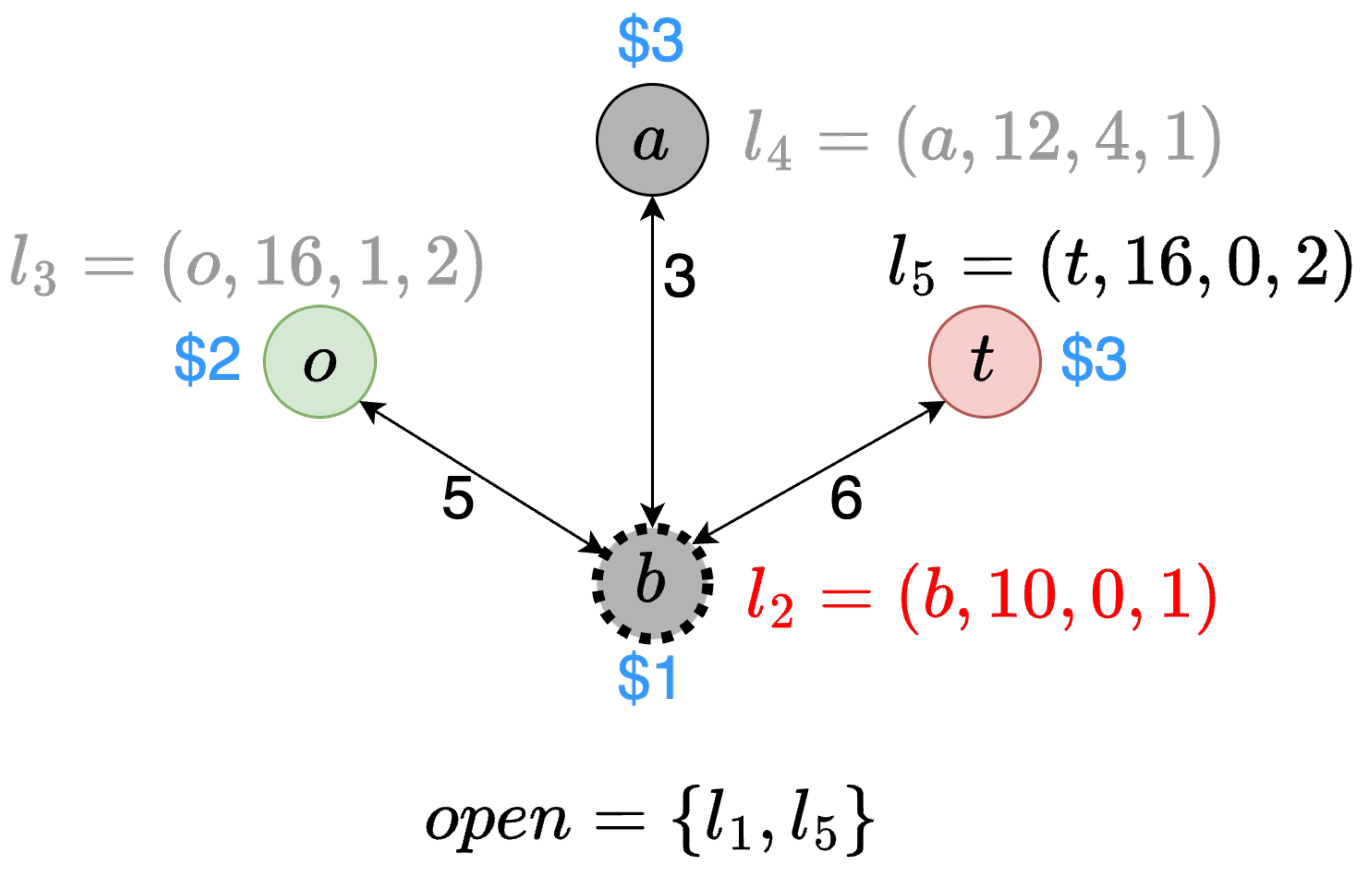}	
		\caption{}
		\label{fig:rf-demo-3}
	\end{subfigure}%
	\begin{subfigure}{.50\linewidth}
		\includegraphics[width=\linewidth]{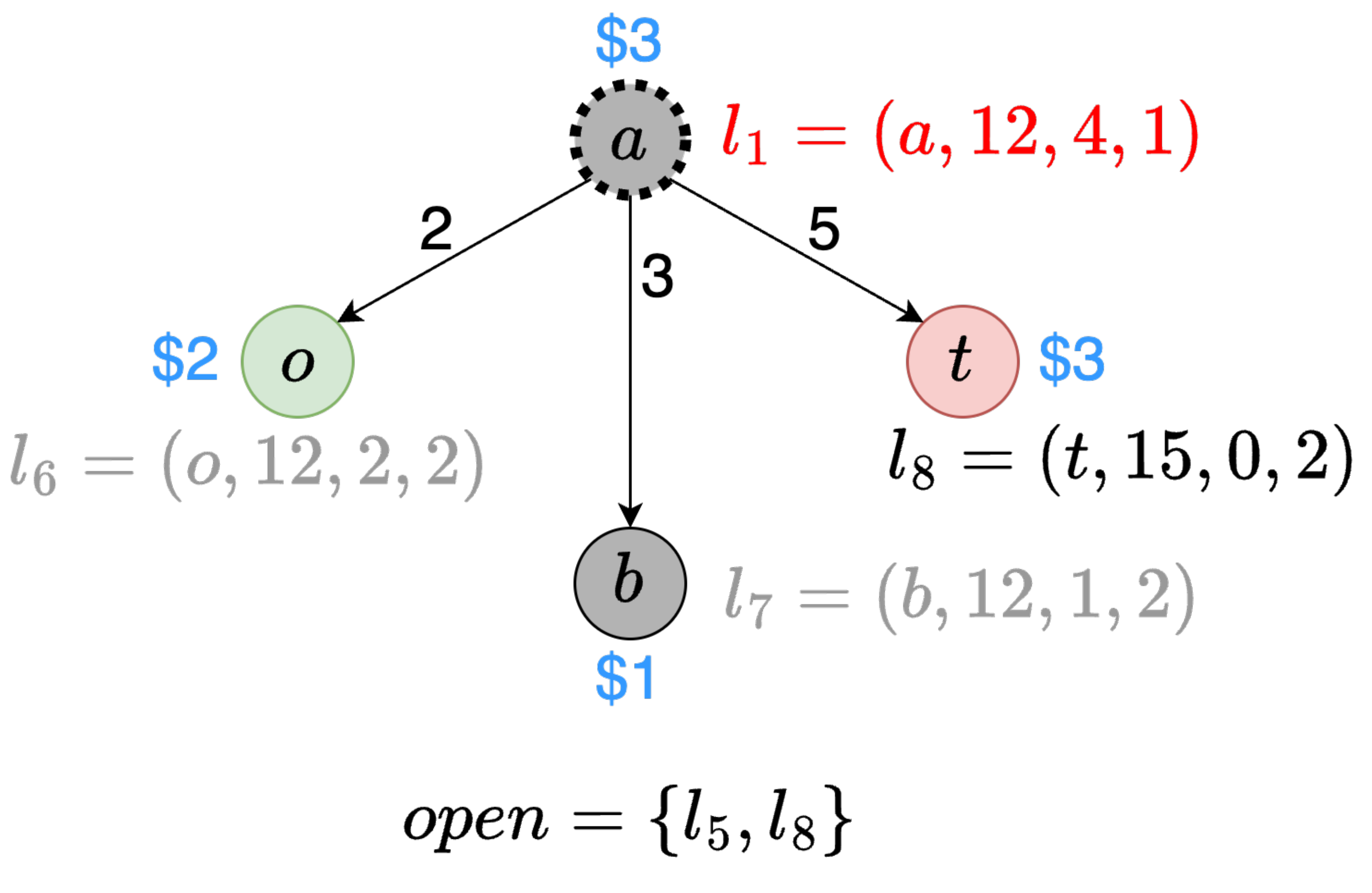}	
		\caption{}
		\label{fig:rf-demo-4}
	\end{subfigure}

  \caption{
  A toy example showing the $\text{Refuel A}^{*}$ with four vertices. 
	The graph $G$ has four vertices $(o,a,b \text{ and } \rev{t})$, with fuel consumption in black, fuel price in blue, $q_{max}=6$ and $k_{max}=2$. 
The i-th label associated with a vertex $v$ is \rev{$l_{i} = (v, g_i, q_i, k_i)$}. 
% (a) In the initial stage, $l_0$ is generated at the starting location $o$.
% (b) As the first iteration progresses, $l_0$ is popped from the OPEN and generates successors $l_1$ and $l_2$. 
% The gray edge is disregarded as the fuel consumption exceeds the tank capacity.
% (c) In the next iteration, $l_2$ is popped, gray labels ($l_3$ and $l_4$) are dominated by previously generated labels 
% ($l_0$ and $l_1$), hence the only remaining successor is $l_5$.
% (d) Similarly, $l_1$ is popped next and generates the undominated successor $l_8$. 
% This results in the OPEN becoming $\{l_8, l_5\}$.
% $l_8$, which is located on the target, becomes the next label to be popped, 
% meaning that the process can be terminated.
The optimal path is shown to be \rev{$o \to b \to t$}, with a cost of $15$.
}
  \label{fig:working}
\end{figure}

\subsection{\abbrRAstar Example}
\label{ppwr:sec:Method:refuel_work}

An example of Alg.~\ref{ppwr:alg:refuelA} \abbrRAstar is shown in Fig.~\ref{fig:working}.
It considers a graph $G$ with four vertices \rev{$o,a,b, \text{and } t$}.
The source is $o$, and goal \rev{$t$}.
The label $l_0 = (o, 0, 0, 0)$ is initiated and inserted into OPEN. 
In the first search iteration, as seen in Fig.~\ref{fig:working}(b), $l_0$ is popped from OPEN. 
$l_0$ does not reach the goal vertex.
Its reachable vertices are computed, which leads to labels $l_1=\{a, 12, 4, 1\}$ and $l_2 = \{b, 10, 0, 1\}$. 
To compute $l_1$, we can see that, since $c(a)>c(o)$, the robot fills up the tank and then moves to $a$ which causes $g(l_1)=12$, $q(l_1)=6-2 =4$ and $k(l_1) = 1$. 
Alternately, $c(b)<c(o)$, so the robot only fills up the amount needed to go to $b$, leading to label $l_2$ with $g(l_2) = 10$, $q(l_2)=5-5 =0$ and $k(l_2) = 1$.
Both $l_1$ and $l_2$ are added to the OPEN as they are non-dominated. 
Next (Fig.~\ref{fig:working}(c)), $l_2$, which has the lowest $f$-value, is popped from OPEN, and follows the same
process for generating labels $l_3, l_4 \text{and } l_5$ corresponding to vertices $o, a$ and \rev{$t$} with $k(l_3) = k(l_4) = k(l_5) = 2$.
Subsequently, $l_3$ and $l_4$ are pruned by $l_0$ and $l_1$ respectively, using $\CheckForPrune$. 
Hence, only $l_5$ is added into the OPEN, that is, OPEN = $\{l_1, l_5\}$. 
In the next iteration, as shown in Fig.~\ref{fig:working} (d), $l_1$ is popped and generates $l_8$. 
The OPEN becomes $\{l_8,l_5\}$. 
$l_8$ is the next popped label. It reaches the goal and represents an optimal solution.
The path and the minimum cost are \rev{$o \to b \to t$} and $15$ respectively.

\subsection{Analysis}
\label{ppwr:sec:Method:analysis}

% In this section, we show why \abbrRAstar is correct as well as provide its run-time complexity.
% We first discuss the run-time complexity of \abbrRAstar and then analyze the completeness and solution optimality.

% \subsubsection{Run-time complexity}
% \label{ppwr:sec:Method:analysis:runtime}
% The dynamic programming (DP) algorithm presented in \cite{khullerFillNotFill2011} has a run-time complexity of $O(k_{max} n^2 \log(n))$, where they provide an $O(n\log(n))$ algorithm to compute one entry corresponding to a vertex and $q > 1$. Since there are $n$ possible choices of vertices and $k_{max}$ possible stops, the overall algorithm stands with an $O(k_{max} n^2 \log(n))$ complexity. The naive method of filling the DP table holds a complexity of $O(k_{max} n^3)$ as explained in ~\cite{khullerFillNotFill2011}.

% A brief analysis and summary of the DP method has been provided in Sec. ~\ref{ppwr:sec:Method:dp_baseline} But, in practice, it is significantly faster than $O(k_{max} n^2 \log(n))$, which is verified in the results section.

In the worst case, \abbrRAstar has the same run-time complexity $O(k_{max} n^3)$ as the basic version of the DP method~\cite{khullerFillNotFill2011}.
This scenario for \abbrRAstar may occur when the heuristic is absent (e.g. $h(v)=0$ for any $v\in V$), dominance pruning does not occur, and all possible labels are expanded.
However, as shown in Sec.~\ref{ppwr:sec:results}, in practice, \abbrRAstar is often much faster than the DP method.
The following theorem summarizes this property.

\begin{theorem}
    \abbrRAstar has polynomial worst-case run-time complexity.
\end{theorem}

% ==========

% 1. a brief review the complexity of Khuller's algorithm, and why.

% 2. Refuel A* in the worst case, still have the same runtime complexity as Khuller's algorithm. 

% 2b. (* construct an example where Refuel A*)

% 3. In practice, Refuel A* is much faster than Khuller's DP, which is verified in the result section.

% =========

% 1. Feasibility. -> infeasible paths are pruned during the search.

% 2. The search is "complete". explore all possible actions of the robot without missing actions (Lemma 1).

% 3. If a label (path) is pruned, then this label cannot lead to an optimal solution. (prove by contradiction. cut-and-paste)

% 4. Heuristic.

% 5. Theorem. based on 1,2,3,4.

% \subsubsection{Completeness and Optimality}
% \label{ppwr:sec:Method:analysis:proofs}

% Now, we demonstrate that our method is correct and provides an optimal path:

In Alg.~\ref{ppwr:alg:refuelA}, due to Line \ref{alg1:line:kmaxCheck}, \abbrRAstar never expands a label $l$ with $k(l)=k_{max}$.
As a result, \abbrRAstar never generates labels with more than $k_{max}$ refuelling stops, and the path returned by \abbrRAstar is feasible, i.e., does not exceed the limit on the number of stops $k_{max}$.
The following lemma thus holds.

\begin{lemma}[Path Feasibility]
\label{ppwr:lemma2}
 The path returned by \abbrRAstar is feasible.
 % The algorithm prunes all paths that cross $k_{max}$ limit.
 \end{lemma}
%  \textit{Proof :}
% A path becomes infeasible when it accumulates stops exceeding $k_{max}$. This emerges due to the expansion of dominated labels. A label is subjected to dominance if its number of stops ($k(l)$) surpasses that of other undominated labels in $\FrontierSet(v(l))$. This implies that more stops were made during the journey from $v_o$ to $v$ compared to the undominated labels associated with vertex $v$. Our algorithm employs the $\CheckForPrune$ function, derived from Definition ~\ref{ppwr:def:dominance}, to eliminate dominated labels. By comparing each label $l$ against all labels $l' \in \FrontierSet(v(l))$, $\CheckForPrune$ evaluates whether the dominance criteria defined in Definition ~\ref{ppwr:def:dominance} are met. If any $l'$ dominates $l$, the label $l$ is pruned; otherwise, its incorporated into $\FrontierSet(v(l))$.

To expand a label, \abbrRAstar considers all reachable neighboring vertices as described in Sec.~\ref{ppwr:sec:Method:compute_reach} and determine the amount of refuelling via Lines \ref{alg1:line:gqk1}-\ref{alg1:line:gqk2}.
With Lemma \ref{ppwr:lemma1}, the expansion of a label $l$ in \abbrRAstar is complete, in a sense that, all possible actions of the robot, which may lead to an optimal solution, are considered during the expansion.
The following lemma summarizes this property.

\begin{lemma}[Complete Expansion]
\label{ppwr:lemma3}
    The expansion of a label in \abbrRAstar is complete.
\end{lemma}

% \textit{Proof :}  
% Following Lemma ~\ref{ppwr:lemma1}, there are 2 possible actions that the robot can take, based on which labels are expanded. Depending on the refuelling cost of vertices along an optimal path, the algorithm ensures that the robot either fills its tank or fills enough to reach only the next vertex. Thus, all actions are explored, and therefore the search is complete. 

During the search, if a label is pruned by dominance in \CheckForPrune, then this label cannot lead to an optimal solution for the following reasons.
If a label $l=(v,g,q,k)$ is dominated by any existing label $l'=(v,g',q',k') \in \FrontierSet(v(l))$.
This means that $l$ has a higher cost $g \geq g'$, lower remaining fuel $q \leq q'$ and has made more stops $k \geq k'$ than $l'$.
Assume that expanding $l$ leads to an optimal solution $\pi_*$, and let $\pi_*(v,v_g)$ denote the sub-path within $\pi_*$ from $v$ to $v_g$.
Then, another path $\pi'$ can be constructed by concatenating the path represented by $l'$ from $v_o$ to $v$, and $\pi_*(v,v_g)$ from $v$ to $v_g$.
Path $\pi'$ is feasible and its cost $g(\pi') \leq g(\pi_*)$.
So, $\pi'$ is a better path than $\pi_*$, which contradicts with the assumption.
We summarize this property with the following lemma.
% $\pi_{1}$, a path to $v$.
% Since we assume that it is an optimal path, it minimizes the refuelling cost, $g(l)$, to $v$. But, we have already mentioned that $l'$ dominated $l$, that is $g(l') < g(l)$. Therefore, our assumption of $g(l)$ being the minimum and hence $\pi_1$ being optimal is wrong. Thus, only non-dominated labels can lead to optimal paths.

\begin{lemma}[Dominance Pruning]  
\label{ppwr:lemma4}
    Any label that is pruned by dominance cannot lead to an optimal solution.
\end{lemma}

We now show that \abbrRAstar is complete and returns an optimal solution for solvable instances.

\begin{theorem}[Completeness]
    For unsolvable instances, \abbrRAstar terminates in finite time. For solvable instance, \abbrRAstar returns a feasible solution in finite time.
\end{theorem}

\IEEEproof{
    Due to Lemma~\ref{ppwr:lemma1} and that the graph $G$ is finite, only a finite number of possible labels can be generated during the search.
    With Lemma~\ref{ppwr:lemma3}, the expansion of a label is complete, which means, \abbrRAstar eventually enumerates all possible labels.
    % For any generated label $l'$, it ends up with one of the following three cases, (i) expanded, (ii) pruned by dominance, or (iii) pruned by $k_{max}$.
    For an unsolvable instance, \abbrRAstar terminates in finite time after enumerating all these labels.
    For solvable instances, due to Lemma~\ref{ppwr:lemma2}, \abbrRAstar terminates in finite time and finds a label that represents a feasible path.
}

\begin{theorem}[Solution Optimality]
\label{ppwr:Theorem1}
For solvable instances, the path returned by \abbrRAstar is an optimal solution.
% and feasible solution path for our \abbrPFRF problem.
\end{theorem}

\IEEEproof{
    When a label $l$ is popped from OPEN and claimed as a solution by \abbrRAstar, due to Lemma~\ref{ppwr:lemma:admissibleHeu}, any other labels in OPEN and their successor labels cannot lead to a cheaper solution than $g(l)$.
    With Lemma~\ref{ppwr:lemma4}, the pruned labels cannot lead to a cheaper solution than $l$.
}

% \input{source/Method/Unbounded_stops}

%%%%%%%%%%%%%%%%%%%%%%%%%%%%%%%%%%%%%%%%%%%%%%%%%%%%%%%%%

% \input{result/MICP.tex}
% \input{Method/khuller_alg_discuss} %discussion on S.Khuller's algorithm
\section{Experimental Results}\label{ppwr:sec:results}

% \shizhe{ToDo:\\
% 	1. Are outliers included in Boxplot median? Yes\\
% 	2. Desc res: overall ..., up to ... \\
% 	3. Show res: RFA*, RFA* exclude heur time, RFA* without heur \\
% 	4. Show res: distance v.s. speed-up, to show effectiveness of heuristic when distance is longer;\\
% }

This section compares the performance of \abbrRAstar with DP~\cite{khullerFillNotFill2011}, and a Mixed-Integer Programming (MIP) model, which are described in Sec.~\ref{sec:baselines}.
Both DP and \abbrRAstar require a preprocessing to compute the reachable set for all vertices in the graph (Alg.~\ref{ppwr:alg:preG}). 
Although this preprocessing is time consuming and takes most of the runtime of both methods, it only needs to run once to support arbitrary number of invocation on the planner, allowing the runtime to be amortized. 
So we exclude the preprocessing time from the runtime, readers can find it from Table~\ref{tab:dataset}.
In the remaining sections, all runtime values are solely for the search after the preprocessing.

\begin{figure}[tb]
	\begin{subfigure}{.45\linewidth}
	\centering
	\includegraphics[width=\linewidth]{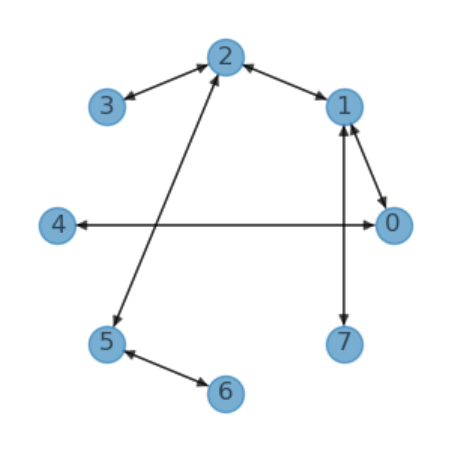}
	\caption{}
	\label{fig:small-graph}
	\end{subfigure}\hfill
	\begin{subfigure}{.45\linewidth}
	\centering
	\includegraphics[width=\linewidth]{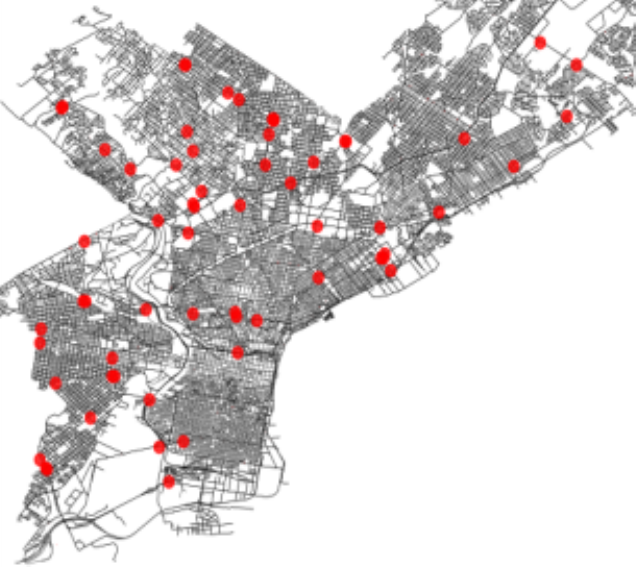}
	\caption{
	}
	\label{fig:city-graph}

	\end{subfigure}
	\caption{
		(\subref{fig:small-graph}) Visualization of a random map.
		(\subref{fig:city-graph})
  Visualization of the map for Philadelphia, USA. 
	It showcases the road network of the city, where the red dots represent the gas stations. 
 %  We create a ``gas station graph'' $G=(V, E)$, where $V$ is the set of gas stations, 
	% and each edge is a minimum fuel consumption path between the corresponding gas stations. 
	% We create this graph $G$ during preprocessing, and both algorithms (\abbrRAstar and DP) runs on $G$ instead of the original city road network. 
	% Blue arrows show the refuelling stops returned by our \abbrRAstar. The agent refuels at 5 stops (including the start location) before reaching the destination.
	}
	\label{fig:expr-graph}
\end{figure}

In the experiments, we use both a synthetic and a real-world dataset.
Fig.~\ref{fig:expr-graph} shows an example of each dataset. 
\rev{The synthetic dataset includes three small (\emph{Synth-S}) and three large (\emph{Synth-L}) random graphs.
Each random graph is a binomial graph that has a single connected component, with a probability of $0.3$ for edge creation~\cite{gilbert1959random}.
%For each pair of different vertices $u,v$ in the graph, the probability that the edge between $u,v$ exists is $0.3$.
The vertex numbers are 8, 16 and 32 for \emph{Synth-S}, and 256, 512, 1024 for \emph{Synth-L}.
For the synthetic maps, the refuelling cost $c$ at vertices are random integers from 1 to 10. 

The real-world dataset consists of five road networks from OpenStreetMap.
For each city map, the refuelling cost $c$ at vertices are randomly sampled from 2.5 to 4.5\footnote{Values are referenced from \url{https://gasprices.aaa.com}}.
Table~\ref{tab:dataset} presents a summary of the graphs.

Table~\ref{tab:default-param} shows the default parameters, where the tank capacity $q_{max}$, and the maximum refuelling stop $k_{max}$ are chosen based
on Table~\ref{tab:dataset} in the following way.
For small maps (\emph{Synth-S}), we set $q_{max}$ and $k_{max}$ to ensure that there is always a solution.
For large maps (\emph{Synth-L} and \emph{City}), we set $q_{max}$ to approximately three times the average edge cost.}
% which results in reachable sets in Algo~\ref{ppwr:alg:preG} having a moderate size.
% }
A pair of start and goal vertex in a map defines a \abbrPFRF instance.
\rev{
For all instances, we confirmed that both DP and \abbrRAstar always have the same cost, which is not surprising since they both guarantee solution optimality.
% Meanwhile, MIP sometimes has a slightly higher cost due to rounding errors.
}
% Def 'task' in Def. 1
Each \abbrPFRF instance has a 30-seconds runtime limit.
All methods are implemented in C++ and tested on Ubuntu 22.0 Desktop with a 13th Gen Intel i7-13700 and 32GB RAM.
\footnote{Our software and dataset is available at \url{https://github.com/rap-lab-org/public_refuelastar}}
% with an academic license.
% \begin{figure}[tb]
% 	\centering
% 	\includegraphics[width=.4\linewidth]{Pics/small-graph.png}
% 	\caption{Visualization of G10.}
% 	\label{fig:small-graph}
% \end{figure}

\addtolength{\tabcolsep}{-1pt}    
\begin{table}[tb]
\centering
\begin{tabular}{llllll}
\toprule
 Dataset       & Map     & $|V|$ & $|E|$  & $E_{median}$  & Preproc.(s) \\
\midrule
% Synthetic maps & 8       & 4     & 13     & 5.0           & 0.0         \\
%                & 16      & 5     & 15     & 5.0           & 0.0         \\
%                & 32      & 10    & 43     & 4.0           & 0.0         \\
%                & 256     & 11    & 111    & 5.0           & 0.0         \\
%                & 512     & 18    & 307    & 5.0           & 0.0         \\
Synth-S        & 8       & 8     & 15     & 5.0           & 0.0			    \\
               & 16      & 16    & 51     & 4.0           & 0.0         \\
               & 32      & 32    & 269    & 4.0           & 0.0         \\
\midrule
Synth-L        & 256     & 256   & 19473  & 4.0           & 0.2         \\
               & 512     & 512   & 78783  & 4.0           & 2.8         \\
               & 1024    & 1024  & 314529 & 4.0           & 33.7        \\
\midrule
 City          & Phil    & 61    & 3661   & 9920.4        & 0.0         \\
               & Austin  & 87    & 7483   & 9842.9        & 0.0         \\
               & Phoenix & 178   & 31507  & 19021.7       & 0.2         \\
               & London  & 258   & 66307  & 22930.7       & 0.7         \\
               & Moscow  & 423   & 178507 & 20706.2       & 3.6         \\
\bottomrule
\end{tabular}
\caption{Summary of the two datasets. $E_{median}$ is the median edge cost. \textit{Preproc.} is the preprocessing time in second.}
\label{tab:dataset}
\end{table}
\addtolength{\tabcolsep}{1pt}    

\begin{table}[tb]
	\centering
	\begin{tabular}{llll}
		\toprule
		Parameters$\backslash$Datasets   & Synth-S & Synth-L & City \\
		\midrule
		$q_{max}$ & 8                       & 15                           & 60000        \\
		$k_{max}$ & 3                       & 10                           & 10          \\
		\bottomrule
	\end{tabular}
	\caption{Default input parameters of each dataset, 
	where \emph{Synth-S} and \emph{Synth-L} are synthetic random maps with 
	small size (8, 16, 32) and large size (256, 512, 1024) respectively} 
	\label{tab:default-param}
\end{table}

\begin{figure}[tb]
	\centering
	\includegraphics[width=.9\linewidth]{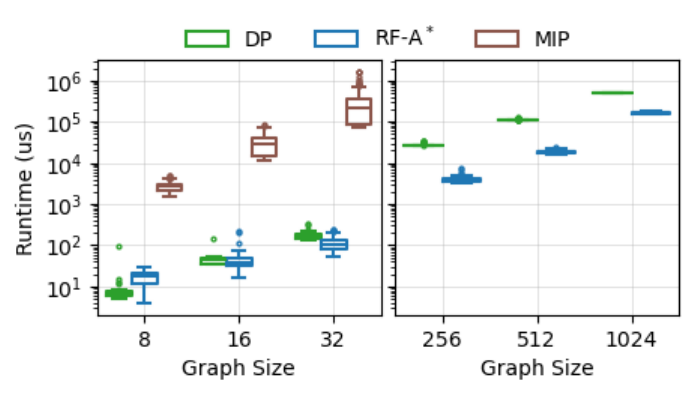}
	\caption{Runtime on the synthetic dataset.}
	\label{fig:small-runtime}
\end{figure}

\begin{figure*}[bth]
	\centering
	\includegraphics[width=\linewidth]{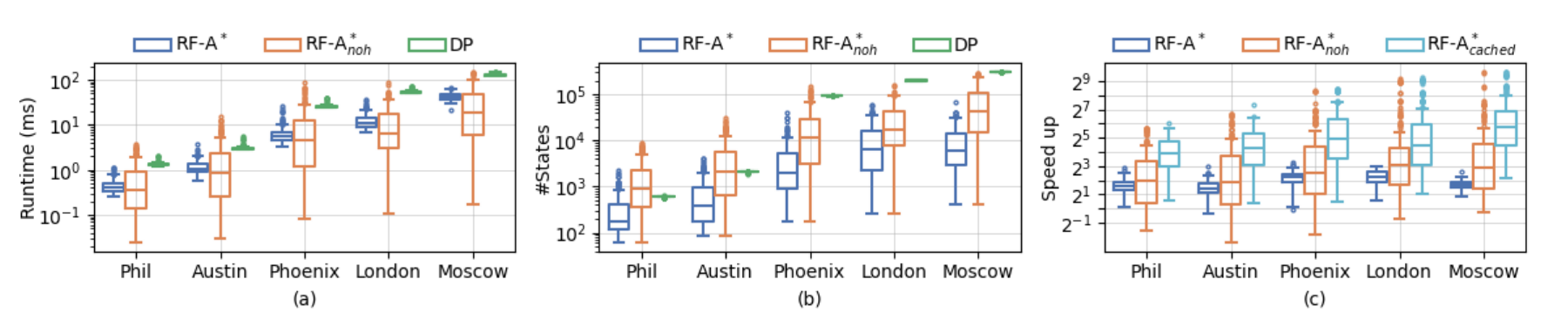}
	\caption{(a) and (b) show runtime and memory cost of all methods. 
	(c) shows the speed-up of \abbrRAstar variants compare to DP.}
	\label{fig:city-res}
\end{figure*}

\begin{figure*}[bth]
	\centering
	\includegraphics[width=\linewidth]{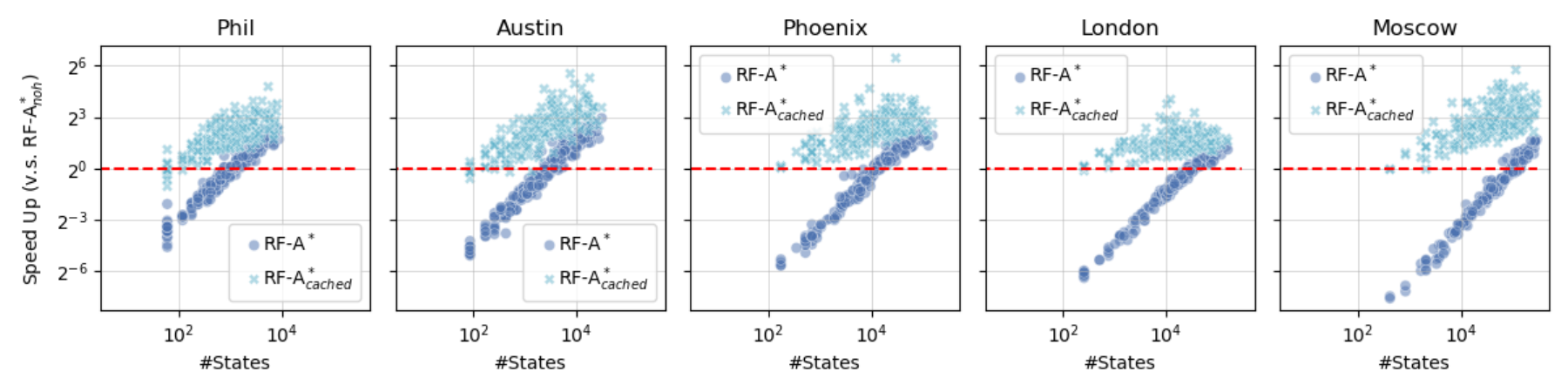}
	\caption{Speed-up compared to \RFDij , with the increase in its explored states. 
	Points below the red dash lines indicate performance worse than \RFDij.}
	\label{fig:size-speedup}
\end{figure*}
\subsection{Baselines}\label{sec:baselines}
% \vspace{-1.5mm}
\subsubsection{\rev{Dynamic Programming (DP)}}\label{sec:dp_baseline}
We provide details of DP for \abbrPFRF from \cite{khullerFillNotFill2011}.
This DP defines a set of sub-problems where each one is $(v,k,q)$ with $ v\in V$, $k$ denoting the number of stops and $q$ denoting the gas level. 
For each sub-problem, let $A(v, k, q)$ represent the minimal cost to traverse from vertex $v$ to the goal $v_g$, within $k$ refuelling stops, and starting with $q$ units of fuel.
With the help of Lemma~\ref{ppwr:lemma1}, for each $v\in V$, there is only a finite number of possible values that $q$ can take, which is bounded by $|V|$.
As a result, the set of sub-problems is finite since each of $v,k,q$ can take a finite number of possible values.
The base case of the DP method is $A(v_o, 0, 0)=0$, and the default value for all other sub-problems are $\infty$.
Then, the DP method iteratively solve all sub-problems $A(v, k, q)$ based on the base case or previously solved sub-problems. 
The optimal solution can be obtained from $A(v_g,k_{max},0)$. In~\cite{khullerFillNotFill2011}, two methods are presented:
a naive method with a time complexity of $O(k_{max}n^{3})$ and an advanced method reduces the complexity to 
$O(k_{max}n^2 \log(n))$.
We use the advanced method for comparison in our experiments.

\subsubsection{\rev{Mixed Integer Programming Formulation}}\label{ppwr:sec:results:mip}
We introduce a simple Mixed Integer Programming formulation as an alternative baseline to solve the \abbrPFRF.
The solver for the MIP model is Gurobi 11.
A binary variable $x(u, v)$ indicates if a path passes through the edge from $u$ to $v$, 
and binary variable $y(u)$ indicates whether the robot refuels at the vertex $u$.
Decision variables $a(u),q(u)$ are introduced in Sec.~\ref{ppwr:sec:problem_state}.
The objective function in~Eq.\ref{res:mip_obj} is the total fuel cost along the path 
(borrowed from Sec.~\ref{ppwr:sec:problem_state}), and is to be minimized.
Eq.~\eqref{res:vars} and Eq.~\eqref{res:mip_y} define the domain of decision variables.
Eq.~\eqref{res:stop_constr} defines the limit on the refuel stops, 
and Eq.~\eqref{res:flow_constr} defines the path from $v_o$ to $v_g$.
Eq.~\eqref{res:gas_conservation} indicates that if an edge $(u, v)$ is on the path, i.e.,
$x(u, v)=1$, then the remaining fuel at $v$ must be equal to the remaining fuel at $u$ plus 
the amount of refuelling at $u$, minus the consumption on the edge. 
Eq.~\eqref{res:smart_refuel} expresses the optimal refuelling strategy in Lemma~\ref{ppwr:lemma1}.

\begin{equation}\label{res:mip_obj}
	\min_{a(u),q(u),x(u,v)}\left(\sum c(u)a(u)\right)
\end{equation}
\begin{equation}\label{res:vars}
\begin{aligned}
	x(u,v) &\in \{0, 1\} \\
	q(u), a(u) &\geq 0 \\
	q(o) & = 0 \\
	q(u)+a(u) &\leq q_{max}\\
\end{aligned}
\end{equation}
\begin{equation} \label{res:mip_y}
		y(u) = 
	\begin{cases} 
	 1, & a(u) > 0 \\ 
	 0, & a(u) = 0
	 \end{cases}\\
\end{equation}
\begin{equation}\label{res:stop_constr}
	\sum_{u\in V} y(u) \leq k_{\max}
\end{equation}
\begin{equation}\label{res:flow_constr}
	\sum_{v\in V} x(u,v) - \sum_{v\in V} x(v,u) = 
 \begin{cases} 
 1, & u=v_o;\\
 -1, & u=v_g; \\
 0, & u\in V/\{v_o,v_g\}
 \end{cases}
\end{equation}
\begin{equation}\label{res:gas_conservation}
(q(u)+a(u) -d(u,v) - q(v))x(u,v) = 0, \;\; (u, v)\; \in E 
\end{equation}
\begin{equation}\label{res:smart_refuel}
	\begin{aligned}
		\forall x(u, v) = 1\;, (u, v)\in E,\;\\
	\begin{cases}
		a(u) = q_{max}, & c(u)<c(v)\\
		a(u) + q(u) \geq d(u, v), & c(u)\geq c(v)\\
	\end{cases}
	\end{aligned}
\end{equation}

\subsection{Synthetic Dataset Results}

For each map from \emph{Synth-S}, we create an instance for each possible pair of start-goal vertices, 
and for each map from \emph{Synth-L}, we randomly select 100 vertex pairs.
Fig.~\ref{fig:small-runtime} presents the results.

In \emph{Synth-S}, we can see that both DP and \abbrRAstar can find optimal solutions within tens of microseconds.
Although DP initially performs slightly faster than \abbrRAstar when handling a small graph, 
this advantage diminishes as the size of the graph increases. Eventually, in \emph{Synth-L} \abbrRAstar becomes faster than DP by
several factors.
Conversely, MIP is slower than both DP and \abbrRAstar by orders of magnitude, and this gap increases as the size of the graph grows. 
% So we conclude that MIP is outperformed by DP and \abbrRAstar, 
% leading us exclude it from the remainder of the experiments.
We therefore remove the MIP from the subsequent experiments.

\subsection{City Dataset Results}

The runtime of \abbrRAstar includes heuristic computation and search process.
The former requires a backward Dijkstra from the goal location, and the computed heuristic can reduce the runtime of the search by generating fewer labels.
In practice, we can reuse the computed heuristic as long as the goal location remains the same, namely, by {caching} the heuristic.
The overall influence of the heuristic on the search process are two-folds.
On the one hand, it can slow down the search due to the overhead on computing the heuristic. 
On the other hand, it can accelerate the search if the speed-up on search outweighs the overhead, or if no overhead exists when the cached heuristic can be used. 
% This depends on the location of starts and goals, and how the cache mechanism has been implemented in the application.

To show the effectiveness of the heuristic in \abbrRAstar, we introduce additional baselines.
The first, \RFDij, conducts a search without a heuristic.
The second, \RFAnoh, excludes the runtime to compute the heuristic before the search starts and only counts the runtime for search,
representing an ideal situation where a cached heuristic is always available.

In this dataset, instances are 100 randomly selected vertex pairs from the graph.
To compare the number of sub-problems have been explored by \abbrRAstar, \RFDij, and DP, we define the term \emph{state}.
The $\#$States for \abbrRAstar and \RFDij refers to the number of labels that are generated during the search. 
For DP, $\#$States refers to the number of $A(v,k,q)$ whose cost values are computed during the DP iterations.
As shown in Fig.~\ref{fig:city-res}(a) and~\ref{fig:city-res}(b), DP has a similar $\#$States and has similar runtime across various instances.
Compared to DP, \abbrRAstar explores fewer states and needs less runtime, while \RFDij needs less runtime but explores more states in the case of small graphs (e.g., Phil and Austin).

We use the word ``the speed-up of an algorithm'' to denote the ratio of the runtime of DP divided by the runtime of that algorithm. 
Fig.~\ref{fig:city-res}(c) reveals that most results of \abbrRAstar have a speed-up between 2 to 8 times, compared to DP.
The results of \RFDij are distributed across a wider range, which means without the guidance of the heuristic, the search may expand many states that are useless to find an optimal solution.
For the median values, \RFDij shows a better performance than \abbrRAstar, particularly with larger cities, e.g., London and Moscow, which suggests that the major contributor to \abbrRAstar's runtime is the heuristic computation, making \abbrRAstar slower than \RFDij for some instances.
Finally, \RFAnoh combines the benefit of having a heuristic and eliminates the overhead of computing a heuristic, which leads to the fastest approach and the speed-up of \RFAnoh rises towards a range of 8 to 64.

To better understand the effectiveness of heuristic across different instances,
we plot the speed-up factor of \abbrRAstar and \RFAnoh, 
compared to \RFDij with the increasing number of generated state of \RFDij, shown in Fig.~\ref{fig:size-speedup}.
We can see that, with the guidance provided by the heuristic, \abbrRAstar runs faster than \RFDij in instances that require a lot of state generation, despite the overhead of computing the heuristic.
This result indicates that the existing pre-processing techniques on a road network 
(e.g.,~\citep{geisberger2012exact,maheo2021customised}) can be potentially applied to reduce the overhead of computing the heuristic to further expedite the search.

% \input{result/result}
% \input{result/Greedy}
% \input{result/MICP}

%%%%%%%%%%%%%%%%%%%%%%%%%%%%%%%%%%%%%%%%%%%%%%%%%%

\section{Conclusion and Future Work}
\label{ppwr:sec:conclusion}

This paper investigates the \abbrPFRF problem introduced in \cite{khullerFillNotFill2011} and develops \abbrRAstar, a fast $\text{A}^{*}$-based algorithm that leverages the heuristic search and dominance pruning rules.
Numerical results verify the advantage of \abbrRAstar over the existing dynamic programming approach.

\rev{The limitation of our approach is that it relies on the assumption that all information about the graph and fuel cost are known in advance and remains unchanged, which may not hold in practice, e.g., the fuel price changes over time. }
For future work, one can introduce an additional time dimension~\cite{ren22mosipp} to \abbrRAstar if the time-vary information is available.
One can also integrate \abbrRAstar into a predict-then-optimize framework~\cite{DBLP:conf/aaai/DemirovicSG0LRC20} if the environment is not fully observable.
Finally, one can also consider the multi-agent version of the problem~\cite{ren23cbssTRO}.

% For future work, one can investigate situations with a time constraint on refuelling, or multi-agent variant of the problem.
% One can also consider using the fast dominance checking techniques in~\cite{renERCANewApproach,ren22emoa} to further expedite the computation when there are a lot of non-dominated labels during the search.

% \FloatBarrier
%%%%%%%%%%%%%%%%%%%%%%%%%%%%%%%%%%%%%%%%%%%%

\bibliographystyle{plain}
\bibliography{Ref}

\end{document}